# CLASSIFYING VARIABLE-LENGTH AUDIO FILES WITH ALL-CONVOLUTIONAL NETWORKS AND MASKED GLOBAL POOLING

*Lars Hertel[1], Huy Phan[1,2], and Alfred Mertins[1]*

[1]Institute for Signal Processing, University of Luebeck, Germany
[2]Graduate School for Computing in Medicine and Life Sciences, University of Luebeck, Germany
Email: {hertel, phan, mertins}@isip.uni-luebeck.de

## ABSTRACT

We trained a deep all-convolutional neural network with masked global pooling to perform single-label classification for acoustic scene classification and multi-label classification for domestic audio tagging in the DCASE-2016 contest. Our network achieved an average accuracy of 84.5 % on the four-fold cross-validation for acoustic scene recognition, compared to the provided baseline of 72.5 %, and an average equal error rate of 0.17 for domestic audio tagging, compared to the baseline of 0.21. The network therefore improves the baselines by a relative amount of 17 % and 19 %, respectively. The network only consists of convolutional layers to extract features from the short-time Fourier transform and one global pooling layer to combine those features. It particularly possesses neither fully-connected layers, besides the fully-connected output layer, nor dropout layers.

## 1. INTRODUCTION

Identifying the location in which a specific audio file was recorded, e.g. a beach or a bus, and moreover understanding the sound sources inside the recording, e.g. speech, is a challenging task for a machine. The complex sound composition of real life audio recordings makes it difficult to obtain representative features for recognition. However, having a machine that understands its environment, e.g. through acoustic events inside the recording, is important for many applications such as security surveillance and context-aware services.

So far, most of the audio-related recognition systems have used hand-crafted features, mainly extracted from the frequency domain of the audio signal, such as Mel frequency cepstral coefficients (MFCC) [1], log-frequency filter banks [2] and time-frequency filters [3]. However, with the rapid advance in computing power, feature learning is becoming more common [4, 5]. In our previous work [6] we compared time-domain and frequency-domain feature learning using deep neural networks and found the latter to be slightly superior. In this work, we will therefore train our convolutional networks on the short-time Fourier transform (STFT) of the audio segments and evaluate their performance both for the acoustic scene classification task and for the domestic audio tagging task of the DCASE-2016[1] contest.

## 2. DATASETS

### 2.1. Acoustic Scene Classification

The acoustic scene classification dataset consists of multiple recordings from 15 different acoustic scenes, which were recorded in distinct locations. The task is to correctly classify the single label, i.e. the location, of every audio file. For each location, stereo audio recordings of 3–5 min were captured and split into 30 s segments. The recordings were sampled at 44.1 kHz and 24 bit. Originally, the dataset contained 1170 audio segments of 30 s. However, due to radio interference from mobile phones and temporary microphone failures, a few segments got partly corrupted and annotation errors were provided. The clean dataset therefore contains some audio files that are shorter than 30 s. For simplicity, we only selected the left-hand channel of each recording when training and testing our networks.

### 2.2. Domestic Audio Tagging

The dataset for domestic audio tagging is based on recordings made in domestic environments. The objective of this task is to perform multi-label classification on 4 s audio segments. In total, there exist seven different classes which can all occur simultaneously in a single audio segment. The training data is provided at sampling rates of 48 kHz in stereo and 16 kHz in mono. However, the final evaluation data is only provided as monophonic recordings sampled at 16 kHz. We therefore only used the monophonic audio data with a sampling rate of 16 kHz when training our networks.

### 2.3. Data Preprocessing

For data preprocessing, we solely computed the STFT and selected the first half of the symmetric magnitude of the complex transform for each time step. We used an asymmetric

---
[1]http://www.cs.tut.fi/sgn/arg/dcase2016/

Hann window as a window function with a window length of 25 ms and a hop size of 15 ms. The STFT was computed using the librosa[2] library. Furthermore, common feature standardization was performed, i.e zero mean and unit variance for each feature across the dataset. Note that we did not downsample the audio files for the acoustic scene recognition task.

## 3. NEURAL NETWORK

### 3.1. Network Architecture

The architecture of our all-convolutional network with masked global pooling is shown in table 1. The network input in layer 0 is a four dimensional tensor $t \in \mathbb{R}^{b \times d \times h \times w}$ with batch size $b$, depth $d$, height $h$ and width $w$. The dimension of $w$ is the number of Fourier coefficients in the STFT, i.e. 552 for acoustic scene classification and 201 for domestic audio tagging, respectively. The dimension of $h$ is not fixed and can vary due to the global pooling layer. During testing, $h$ is the total number of time steps for each preprocessed audio file, i.e. we input the full audio segment. During training, the size of $h$ depends on the task. For acoustic scene classification, $h$ was randomly chosen within its valid range since the acoustic scene must be recognized independent of the duration of the recording. For domestic audio tagging, $h$ was again the complete segment since we do not know when a certain class occurs.

The first convolution layer 1 performs a one dimensional convolution. It convolves $t$ only in time direction, i.e. in $h$. The filter size is therefore identical with $w$, resulting in $w = 1$ after the convolution. Our network learns 256 filters in layer 1. They can be interpreted as a learned filter bank. The following convolution layer 2, 3 and 4 then combine adjacent time steps. We selected a filter size of 3 and a stride of 2 for all three layers. Note that by using a stride larger than 1, we reduce the size of $h$ and therefore replace the pooling layer, which is conventionally used for this purpose. After each convolution we calculated the nonlinear activation with the common rectified linear unit (relu) [7, 8].

The next layer 5 is a global pooling layer, i.e. we pool the output of all trailing dimensions beyond the second, resulting in a 256 dimensional feature vector $f$. We selected mean-pooling as the pooling function, but virtually any aggregation function could be used. Due to the fact of different lengths of the audio files for acoustic scene classification, we had to adapt the pooling layer slightly for this task. When training the network with mini-batches, an input $t$ of fixed size must be used, i.e. $h$ may not vary. Simply zero-padding the shorter audio signals would distort the results since the additional zeros would be taken into account by the network. We therefore additionally input a mask $m$ to our network. It is a vector of length $b$ that denotes the actual length of each sample in the mini-batch. Our masked global pooling layer

[2] http://github.com/librosa

Table 1: Architecture of our all-convolutional network with masked global pooling.

| No. | Layer | Dimension | | | Parameters |
|---|---|---|---|---|---|
| | | Depth | Height | Width | - |
| 0 | Input | 1 | $h$ | 552 | |
| | | | | (201) | - |
| 1 | Convolution | 256 | $h$ | 1 | 141,312 |
| 2 | Convolution | 256 | $h$ | 1 | 196,608 |
| 3 | Convolution | 256 | $h$ | 1 | 196,608 |
| 4 | Convolution | 256 | $h$ | 1 | 196,608 |
| 5 | Global Pooling | 1 | 1 | 256 | - |
| 6 | Fully Connected | 1 | 1 | 15 | 3,855 |
| | | | | (7) | (2,056) |

then does not take the additionally zero-padded values into account.

The output layer 6 is a standard fully-connected layer that combines each value of our feature vector $f$. It has one neuron for each class, i.e. 15 neurons for acoustic scene classification and 7 neurons for domestic audio tagging. As a nonlinear activation function we selected the softmax function for acoustic scene classification and the sigmoid function for domestic audio tagging. The former calculates a probability distribution over all classes for single-label classification, while the latter outputs a posterior probability for each class for multi-label classification. Note that our proposed network architecture does not use any fully-connected layers, besides the output layer, or dropout [9] layers to regularize the network.

### 3.2. Network Training

To train our networks, we used the Adam [10] gradient descent algorithm with a mini-batch size of 96. The objective function was the multinomial crossentropy for acoustic scene classification and the binary crossentropy for domestic audio tagging. The learning rate starts at 0.001 and is divided by 2 whenever the error plateaus. Early-stopping was used as soon as either overfitting or no meaningful improvement over multiple epochs were recognized. Batch-normalization [11] is applied right after each convolution layer before its nonlinear activation, following Ioffe et al. [11]. This tremendously reduced the necessary training time. We initialized the weights as described by He et al. [12] and trained all networks from scratch. To regularize the networks, we used weight decay of 0.0004.

Table 2: Results for acoustic scene classification.

| Acoustic Scene | Accuracy (%) | |
|---|---|---|
| | Baseline | Network |
| Beach | 69.3 | **78.2** |
| Bus | 79.6 | **83.3** |
| Cafe / Restaurant | 83.2 | 73.1 |
| Car | 87.2 | **91.0** |
| City Center | 85.5 | **96.2** |
| Forest Path | 81.0 | **100.0** |
| Grocery Store | 65.0 | **79.5** |
| Home | 82.1 | **89.7** |
| Library | 50.4 | **91.0** |
| Metro Station | 94.7 | **100.0** |
| Office | 98.6 | **100.0** |
| Park | 13.9 | **53.8** |
| Residential Area | 77.7 | 76.9 |
| Train | 33.6 | **62.8** |
| Tram | 85.4 | **92.3** |
| Average | 72.5 | **84.5** |

Table 3: Results for domestic audio tagging.

| Audio Tag | Equal Error Rate (EER) | |
|---|---|---|
| | Baseline | Network |
| Adult Female Speech | 0.29 | **0.18** |
| Adult Male Speech | 0.30 | **0.20** |
| Broadband Noise | 0.09 | 0.23 |
| Child Speech | 0.20 | **0.06** |
| Other | 0.29 | **0.19** |
| Percussive Sound | 0.25 | **0.11** |
| Video Game / TV | 0.07 | 0.24 |
| Average | 0.21 | **0.17** |

## 4. RESULTS

### 4.1. Acoustic Scene Classification

Our results for acoustic scene classification are given in table 2. It shows the average accuracy in percent across all four folds of the cross-validation for each individual class. For comparison, the baseline [13] results are also given. The classifier of the baseline is a Gaussian mixture model (GMM) with 16 Gaussians per class. It uses 20 MFCC static coefficients, 20 delta coefficients, and 20 acceleration coefficients as features, extracted with a frame size of $40\,\text{ms}$ and $20\,\text{ms}$ hop size. Overall, our network achieves an accuracy of $84.5\,\%$, compared to the average baseline of $72.5\,\%$. This is an absolute improvement of $12\,\%$. There exist only two acoustic scenes, namely *cafe / restaurant* and *residential area*, for which the baseline achieves better results. Three acoustic scenes, namely *forest path*, *metro station* and *office*, are even always correctly classified. The most difficult scene for our network to recognize is the *park*, which coincides with the baseline.

### 4.2. Domestic Audio Tagging

Our results for domestic audio tagging are given in table 3. It shows the equal error rate [14] (EER) averaged across all five folds of the cross-validation for each individual class. For comparison, the baseline [15] results are also given. The classifier of the baseline is a Gaussian mixture model (GMM) with eight Gaussians per class. It uses 14 MFCC static coefficients as features, extracted with a frame size of $20\,\text{ms}$ and $10\,\text{ms}$ hop size. Overall, our network achieves an error rate of $0.17$, compared to the average baseline of $0.21$. This is a relative improvement of $19\,\%$. The easiest audio tag for our network to recognize is *child speech*, whereas *adult male speech* is the most difficult audio tag. Interestingly, our network has problems to label the classes that are most easily recognized by the baseline, i.e. *broadband noise* and *video game / tv*.

## 5. DISCUSSION

We selected a window size of $25\,\text{ms}$ and a hop size of $15\,\text{ms}$ for two reasons. First, window sizes around $25\,\text{ms}$ have been proven useful features for many classifiers [13], including neural networks [6]. Secondly, the size of the window corresponds with the input of our network. Particularly when using the full sampling rate of $44.1\,\text{kHz}$ for acoustic scene classification, increasing the window size would quickly lead to memory constraints and long training times of the network.

We intentionally did not use either dropout or fully-connected layers in our network. Recent research [16, 17] in computer vision showed that convolutional layers alone are strong regularizers. Omitting the fully-connected layers greatly reduces the number of network parameters to learn, which additionally speeds up training.

We observed that maintaining the full sampling rate of $44.1\,\text{kHz}$ was crucial for the achieved accuracy in the acoustic scene classification task. When we downsampled the audio files to $16\,\text{kHz}$, we noticed an overall loss of nearly $10\,\%$. We assume that the equal error rate for domestic audio tagging would also decrease when training the network on the full sampling rate. However, we did not test this since the evaluation data is given only with a sampling frequency of $16\,\text{kHz}$.

Due to memory constraints, we solely used the left-hand channel when training our networks. Using the provided stereo audio files could therefore improve the results. Furthermore, using hand-crafted features, e.g. cepstral coefficients or filter banks besides the learned features might also be useful.

## 6. CONCLUSIONS

We proposed a deep all-convolutional neural network architecture to perform single-label classification for acoustic scene classification and multi-label classification for domestic audio tagging in the DCASE-2016 contest. To handle the different lengths of the audio samples, we introduced a masked global pooling layer. The networks were trained on the short-time Fourier transform of the input signal. They achieved an overall accuracy of $84.5\,\%$ compared to the given baseline of $72.5\,\%$ for acoustic scene classification and an equal error rate of $0.17$ compared to the given baseline of $0.21$ for domestic audio tagging, respectively. Further improvements might be obtained by adding hand-crafted features to the learned feature set.